\documentclass[letterpaper, 10 pt, conference]{ieeeconf}  % Comment this line out if you need a4paper

\IEEEoverridecommandlockouts                              % This command is only needed if 
\usepackage{amssymb}  % assumes amsmath package installed
\usepackage{amsmath}
\usepackage{cite}
 \usepackage{siunitx}
\usepackage{pdfpages}
\usepackage{graphics}
\usepackage{lipsum}
\usepackage{mathtools}
\usepackage{cuted}
\usepackage[T1]{fontenc}
\usepackage[utf8]{luainputenc}
\usepackage{amsbsy}
\usepackage{algorithm}
\usepackage{algpseudocode}
\usepackage{algpascal}
\usepackage{bm}
\usepackage{hyperref}
\usepackage[font=small,skip=0pt]{caption}
\usepackage{graphicx}
\title{\LARGE \bf
Dense-Joint-Based Obstacle-Aided Locomotion with a Joint-Repositionable Snake Robot
}

\author{ Kyosuke Minomo$^{1}$, Ryo Takahashi$^{2}$, Kotaro Yasui$^{3}$, Yasutaka Nakashima$^{1}$, Motoji Yamamoto$^{1}$, and Ayato Kanada$^{4}$ % <-this % stops a space
\thanks{This work was supported by JST PRESTO Grant Number JPMJPR2515, JSPS KAKENHI Grant Number 26K22451, and the Tateisi Science and Technology Foundation.}% 
\thanks{$^{1}$K. Minomo, Y. Nakashima, and M. Yamamoto are with the Department of Mechanical Engineering,
        Kyushu University, Fukuoka 819-0395, Japan
        {\tt\small minomo.kyosuke.366@s.kyushu-u.ac.jp}}%
\thanks{$^{2}$R. Takahashi is with the Graduate School of Engineering, 
        The University of Tokyo, Tokyo 113-8656, Japan
        {\tt\small takahashi@akg.t.u-tokyo.ac.jp}}%
\thanks{$^{3}$K. Yasui is with the Frontier Research Institute for Interdisciplinary Sciences, Tohoku University, Sendai 980-8577, Japan
        {\tt\small k.yasui@riec.tohoku.ac.jp}}%
\thanks{$^{4}$A. Kanada is with the Graduate School of Informatics and Engineering, The University of Electro-Communications, Chofu, Tokyo 182-8585, Japan
        {\tt\small kanada@uec.ac.jp}}%
}

\begin{document}

\maketitle

%%%%%%%%%%%%%%%%%%%%%%%%%%%%%%%%%%%%%%%%%%%%%%%%%%%%%%%%%%%%%%%%%%%%%%%%%%%%%%%%

\begin{abstract} 
Obstacle-aided locomotion is a fundamental capability for snake robots to traverse complex environments. However, conventional rigid-link snake robots often suffer from stagnation or jamming caused by their low joint density (i.e., the number of joints per unit length). This results in discontinuous contact with obstacles, unlike the continuous adaptation of biological snakes. To investigate the effect of joint density on obstacle-aided locomotion performance, we utilized a joint-repositionable snake robot mechanism that decouples actuators from joints, enabling a high-density architecture. We developed two experimental models with identical total lengths but different joint densities (high-density and low-density) and conducted comparative propulsion experiments in obstacle environments with varying obstacle diameters. The experimental results demonstrate that the high-density model substantially suppresses the abrupt shifts in reaction forces that cause stagnation in the low-density model. By maintaining smooth contact points, the high-density configuration reduces power consumption and achieves stable, continuous propulsion. These results highlight high joint density as a key factor in improving the environmental adaptability of snake robots in complex terrains. \end{abstract}

\section{INTRODUCTION}
Biological snakes can push against obstacles to move through complex spaces, a behavior termed \textit{obstacle-aided locomotion} (OAL) that relies on their highly articulated, limbless bodies~\cite{transeth2008, gray1946}.
Equipping snake-inspired robots with such dense joint structures holds the potential for search-and-rescue tasks in complex terrains, unlike conventional wheeled or legged robots which are physically blocked by narrow or irregular environments~\cite{chavan2015, mexico2017}.
However, a fundamental gap remains in joint density (i.e., joint number per length) between biological snakes and snake robots.
While biological snakes have hundreds of vertebrae to maintain continuous contact with their surroundings~\cite{seeja2022}, typical snake robots rely on only tens of joints~\cite{mori2002, wright2012}.
This low joint density requires the snake robots to operate in coarse bending shapes that interact discontinuously with obstacles~\cite{hatton2010}, typically suffering from geometric locking or jamming, leading to unstable propulsion and high power loss~\cite{vespignani2015}.

\begin{figure}[t!]
  \centering
  \includegraphics[width=0.95\linewidth]{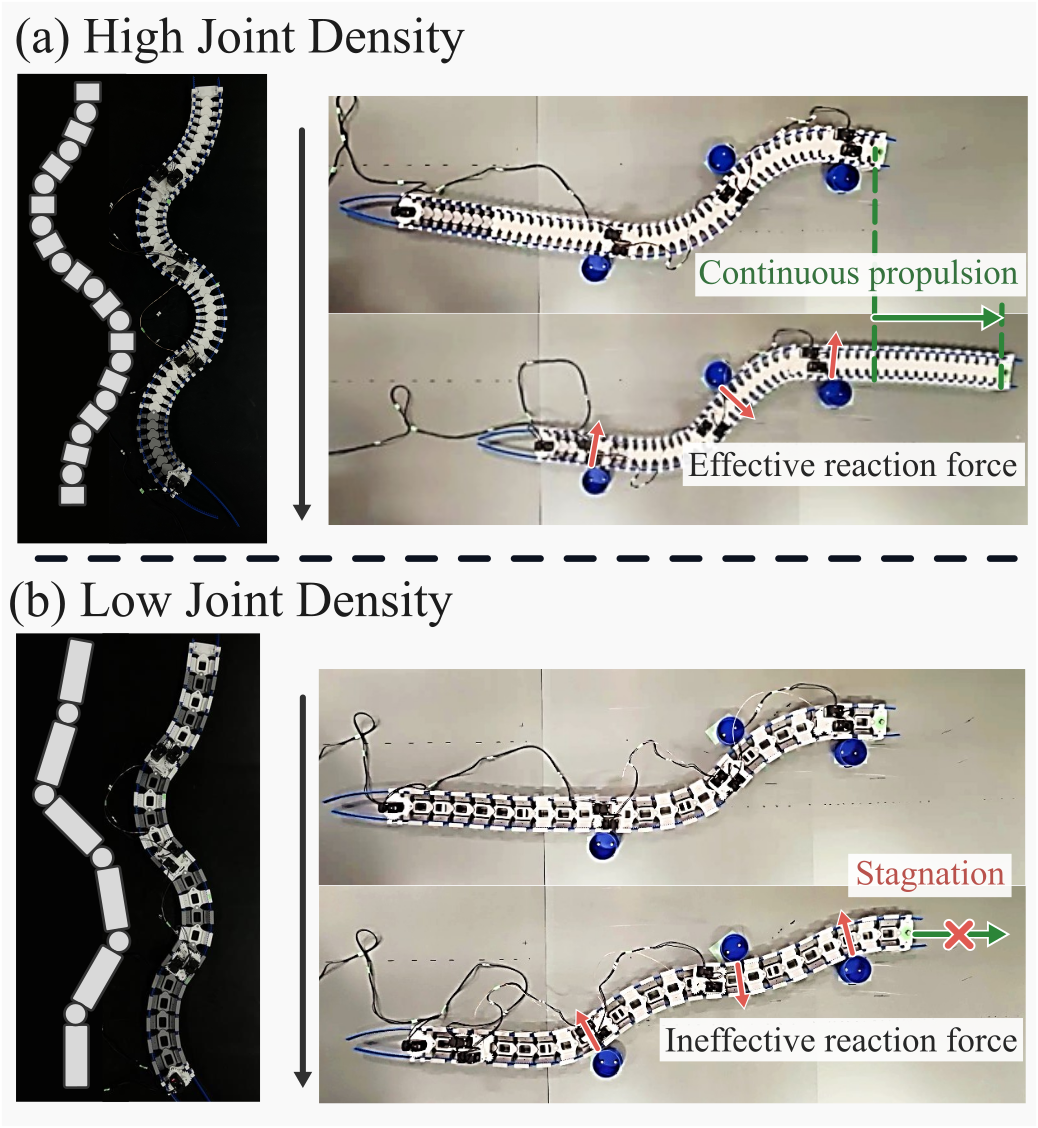}
  \caption{Conceptual comparison of obstacle-aided locomotion based on joint density. (a) The high-density configuration smoothly conforms to obstacles, generating effective reaction forces for continuous propulsion. (b) Conversely, the conventional low-density configuration causes coarse bending and discontinuous contact, leading to abrupt shifts in reaction forces and stagnation.}
  \label{fig:concept}
\end{figure}

To address these challenges, prior studies have improved OAL through active contact feedback~\cite{Fu_2023} or passive mechanical compliance using soft materials~\cite{wang2023}.
The underlying principle shared by these approaches is the stabilization of reaction forces through smoother interactions with obstacles.
However, the specific contribution of \textit{joint density} has remained difficult to isolate because conventional snake robots are constrained by a strict trade-off between actuator size, torque, and the number of joints.
While representative underactuated systems, including wire-driven and soft robots, realize high-density deformation with fewer motors~\cite{racioppo2019, ohno2001, onal2013}, they typically rely on predefined mechanical couplings.
This prevents independent control of local curvature, making it challenging to seamlessly and accurately adjust local body segments to various contact conditions.

The present study focuses on a different physical mechanism: the role of joint density in maintaining continuous body--environment contact.
We utilize a new class of underactuated snake robot named joint-repositionable snake robot, which we previously introduced in~\cite{kanada2025}, and which overcomes the aforementioned limitation by decoupling actuators from joints.
This structure incorporates internally translatable actuator units traveling along flexible rack gears, enabling a high-density articulated body while preserving independent local shape control.
While~\cite{kanada2025} presented the mechanism itself, its functional implications for OAL stability have remained unexplored.
Using this platform, we experimentally compared robots with identical overall dimensions but different joint densities (high-density and low-density), thereby directly evaluating the effect of body discretization on OAL performance in obstacle environments with varying obstacle diameters.

The experimental results demonstrate that increasing joint density enables the smooth \textit{posterior propagation} of body waves~\cite{inazawa2021, takemori2018}, suppressing abrupt transitions of contact points and reaction forces that cause stagnation in conventional sparse-joint designs~\cite{takanashi2023, muller2020} (Fig.~\ref{fig:concept}).
This leads to more stable propulsion and lower energy consumption.
Importantly, although the robot employs flexible rack gears, the observed improvement is primarily attributed to the finer spatial discretization of the body curve rather than compliance itself.
In other words, the high-density configuration enables the body to approximate obstacle geometries more continuously.
Therefore, rather than proposing an alternative to compliance-based approaches, this work identifies joint density as an independent mechanical factor that contributes to continuous body--environment coupling.
These findings clarify the role of body discretization in OAL and provide a complementary design perspective alongside existing control-based and material-based strategies.

%%%%%%%%%%%%%%%%%%%%%%%%%%%%%%%%%%%%%%%%%%%%%%%%%%%%%%%%%%%%%%%%%%%%%%%
\section{ROBOT MECHANISM}
\subsection{Joint-Repositionable System}

\begin{figure}[t!]
  \centering
  \includegraphics[width=0.95\linewidth]{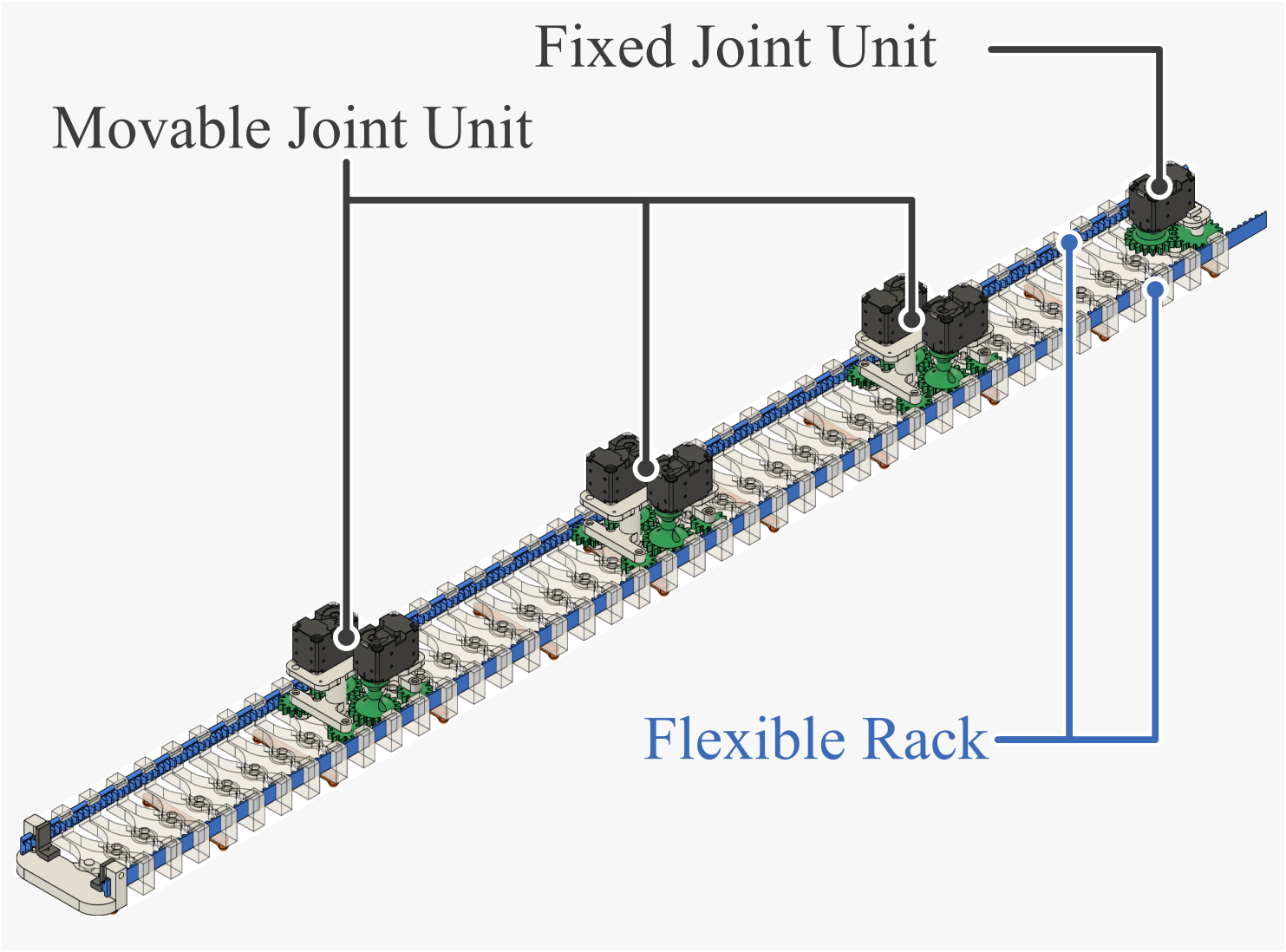}
  \caption{Overall structure of the developed robot.}
  \label{fig:assembly}
\end{figure}

\begin{table}[t]
  \centering
  \caption{Specifications of the experimental snake robot}
  \label{tab:robot_spec}
  \begin{tabular}{lcc}
    \hline
    Item & Dimension [mm] \\
    \hline \hline
    Total Length & 1500 \\
    Total Width & 100 \\
    Total Height & 39.5 \\
    \hline
    Joint Unit Length & 86 \\
    Joint Unit Width & 72 \\
    Active Gear Pitch Radius & 18 \\
    Passive Gear Pitch Radius & 12 \\
    \hline
  \end{tabular}
\end{table}

\begin{figure}[t!]
  \centering
  \includegraphics[width=0.95\linewidth]{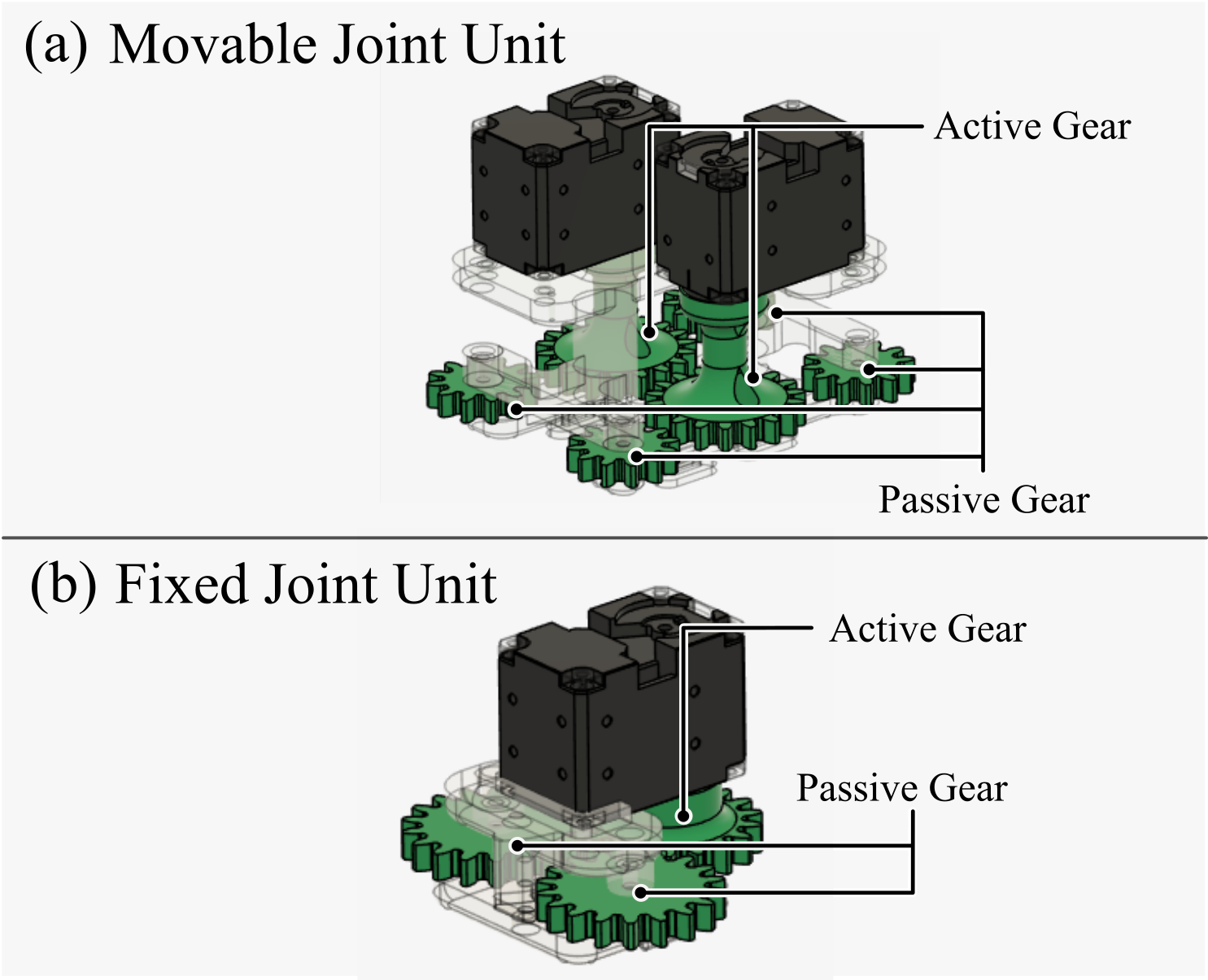}
  \caption{Details of the joint units: (a) Movable joint unit and (b) Fixed joint unit.}
  \label{fig:motor_unit}
\end{figure}

\begin{figure}[t!]
  \centering
  \includegraphics[width=0.95\linewidth]{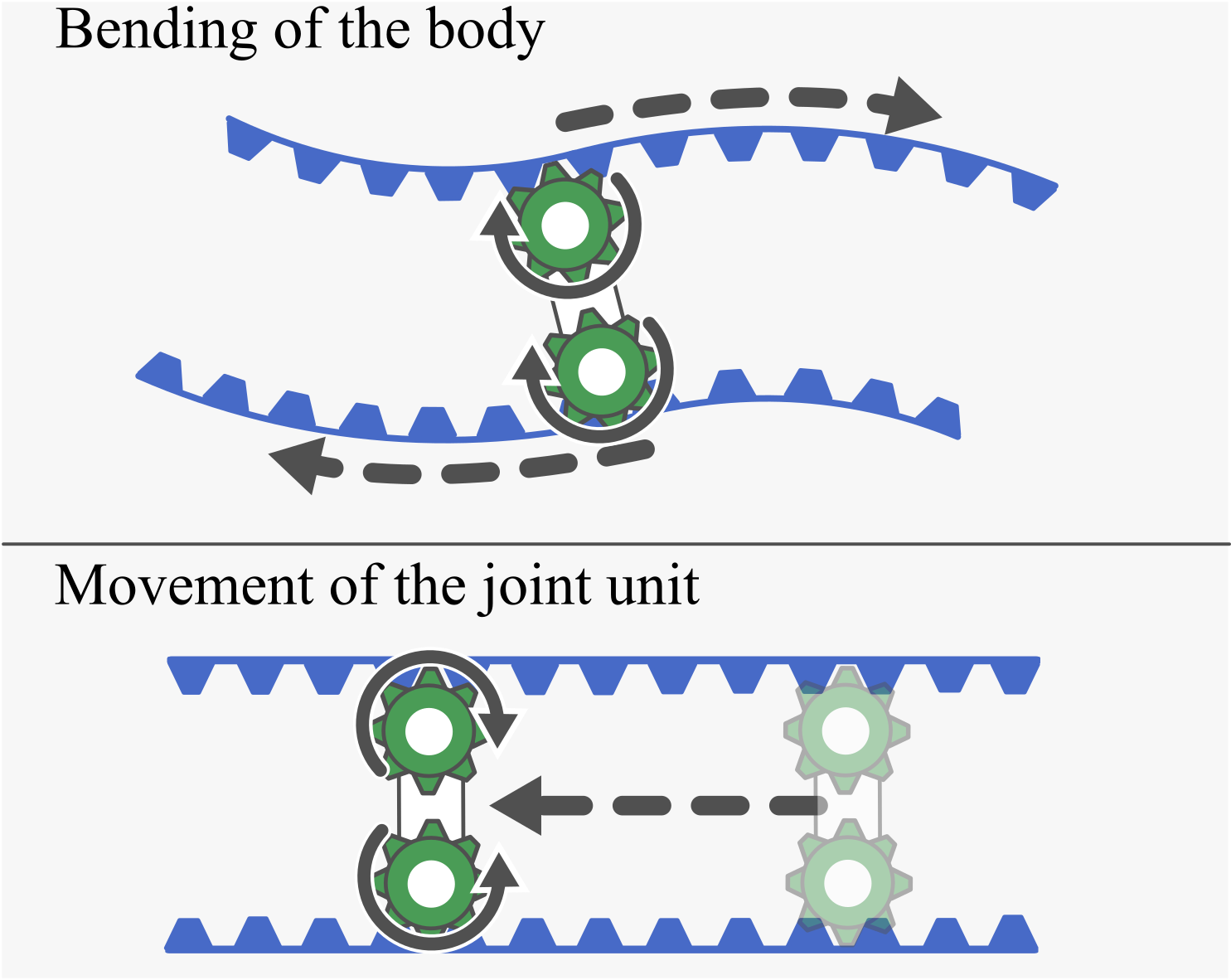}
  \caption{Two distinct operations realized by the joint unit.}
  \label{fig:drive_principle}
\end{figure}

The developed joint-repositionable snake robot consists of (i) two types of joint units (movable joint unit and fixed joint unit) mounted on flexible racks, as shown in Fig.~\ref{fig:assembly} and Table~\ref{tab:robot_spec}, and (ii) a series of passive links connected by rotational joints and two flexible rack gears (metric module 2.0) passing through them. 
Each movable joint unit is equipped with two servomotors (Fig.~\ref{fig:motor_unit}(a)). 
Each motor couples with the rack gears via pinion gears (active gears), and four additional passive gears are arranged for posture stabilization.
By controlling the rotation direction of the two motors, the movable joint unit enables two types of operations (Fig.~\ref{fig:drive_principle}).
The first operation involves the movement of the joint unit inside the robot. 
By rotating the motors in opposite directions, the movable joint unit travels back and forth inside the body along the rack gears, allowing the robot to dynamically change the actuator position.
The second operation generates the bending of the robot's body. 
When rotating the left and right motors in the same direction, one rack gear is pushed forward while the other is pulled backward. 
Since the tips of the rack gears are fixed, the length difference on the two sides can generate a bending moment in the body, creating an S-shaped bending centered at the joint unit's position.
Furthermore, to continuously define the arc shape of the trailing body segment during locomotion, a joint unit must be constantly positioned at the tail. 
Since this specific unit does not need to travel along the body, a fixed joint unit (Fig.~\ref{fig:motor_unit}(b)) is placed at the rear end of the robot. 

Unlike the movable joint units, this unit lacks the translational mechanism and uses a single motor to drive the two rack gears, performing only the bending operation of the rear section. 
By combining these movable and fixed units, this mechanism decouples the number of joints from the number of actuators, enabling a high-density joint structure that is mechanically independent.

\subsection{Design for Joint Density Comparison}
\begin{figure}[t!]
  \centering
  \includegraphics[width=0.9\linewidth]{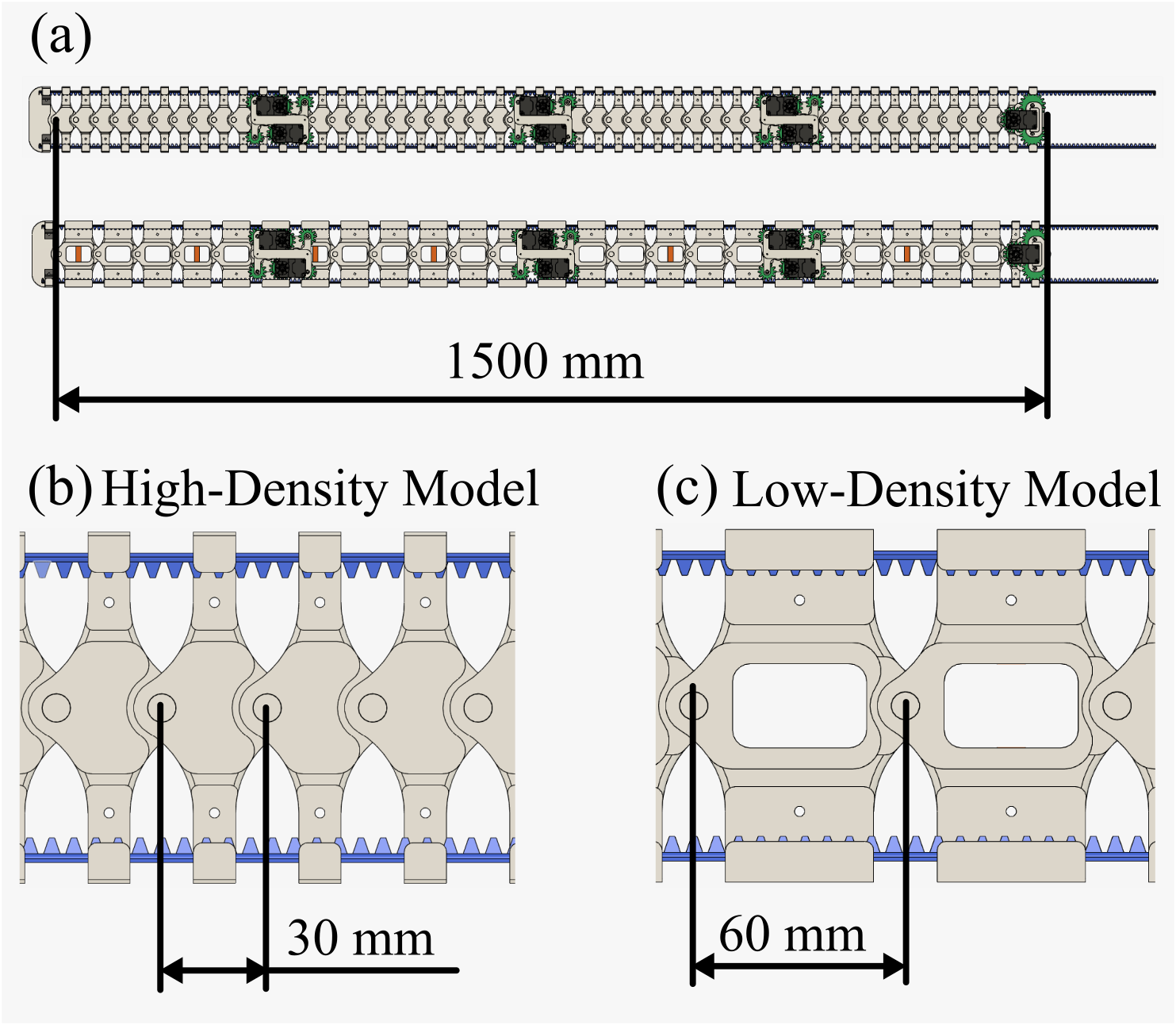}
  \caption{Comparison of High-Density Model and Low-Density Model: (a) Overview of the two models, (b) High-Density Model, and (c) Low-Density Model.}
  \label{fig:two_type_robots}
\end{figure}

To verify the hypothesis that joint density affects obstacle-aided locomotion performance, we developed two experimental models (High-Density Model and Low-Density Model) with different link lengths but identical total body length ($1500$~mm) (Fig.~\ref{fig:two_type_robots}(a)).

The High-Density Model consists of 48 links with a link length of $30$~mm, whereas the Low-Density Model consists of 24 links with a link length of $60$~mm (Fig.~\ref{fig:two_type_robots}(b) and (c)). Note that both models include two common dedicated links with a length of $30$~mm at the rear end for installing the fixed joint unit, ensuring that the total length is accurately unified to $1500$~mm. As a result, the High-Density Model has twice the joint density of the Low-Density Model.

\begin{figure}[t!]
  \centering
  \includegraphics[width=0.9\linewidth]{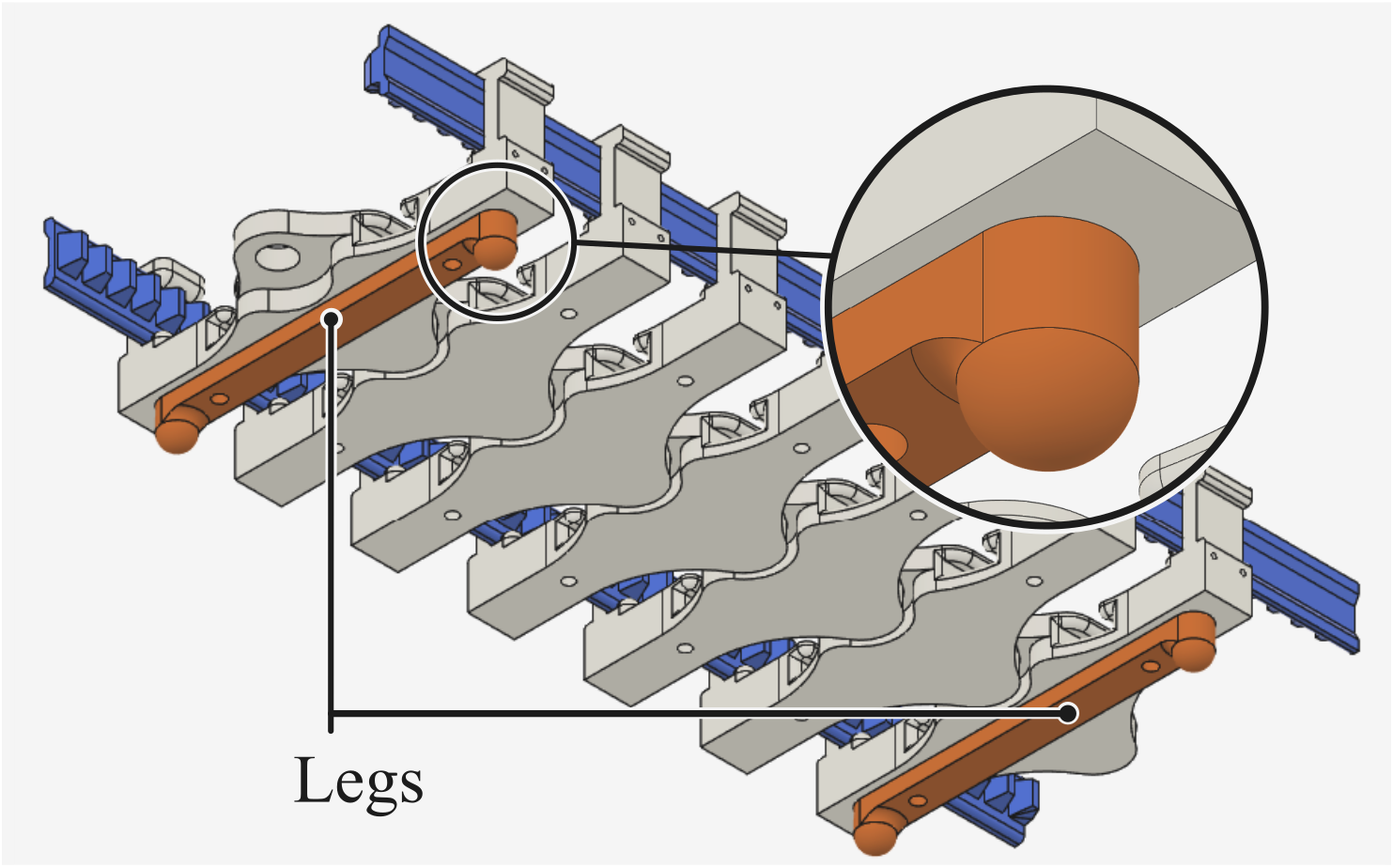}
  \caption{Detail of the spherical-tipped legs attached to the bottom of the links. The spherical shape reduces friction with the ground surface, providing isotropic and low-friction characteristics.}
  \label{fig:legs}
\end{figure}

Typically, snake robots propel themselves using frictional anisotropy provided by wheels or similar mechanisms. However, this study verifies a propulsion principle that utilizes only reaction forces from obstacles. Therefore, spherical legs were attached to the bottom of the links to set the friction characteristics isotropic and low-friction (Fig.~\ref{fig:legs})~\cite{Alben2019, sanfilippo2017}. This eliminates the contribution of propulsion force derived from floor friction.

%%%%%%%%%%%%%%%%%%%%%%%%%%%%%%%%%%%%%%%%%%%%%%%%%%%%%%%%%%%%%%%%%%%%%%%%%%%%%%%%
\section{OBSTACLE-AIDED LOCOMOTION WITH THE JOINT-REPOSITIONABLE SNAKE ROBOT}
In this section, we describe the specific motion generation method for obstacle-aided locomotion using the fabricated experimental robot.
\subsection{Propulsion Principle and Contact Conditions}
To gain propulsion using obstacles, the robot must exert appropriate forces on objects in the environment and convert the reaction forces into thrust in the direction of travel. According to Transeth et al., efficient propulsion using reaction forces from obstacles requires at least three contact points (Push Points)~\cite{transeth2008}.
If there are two or fewer contact points, the robot cannot cancel the moment generated by the reaction forces from the obstacles, causing problems such as the robot rotating on the spot or failing to move in the intended direction.
Therefore, in this experiment, the basic condition for operation was set to maintain simultaneous contact with three obstacles arranged alternately on the left and right relative to the robot's direction of travel. This setup aims to synthesize the reaction forces from each contact point to obtain stable propulsion while suppressing sideslip and rotation.

\subsection{Motion Generation with Joint Units}
Obstacle-aided locomotion in this robot is realized by utilizing the translational capability of the movable joint units.

\begin{figure}[t]
  \centering
  \includegraphics[width=0.9\linewidth]{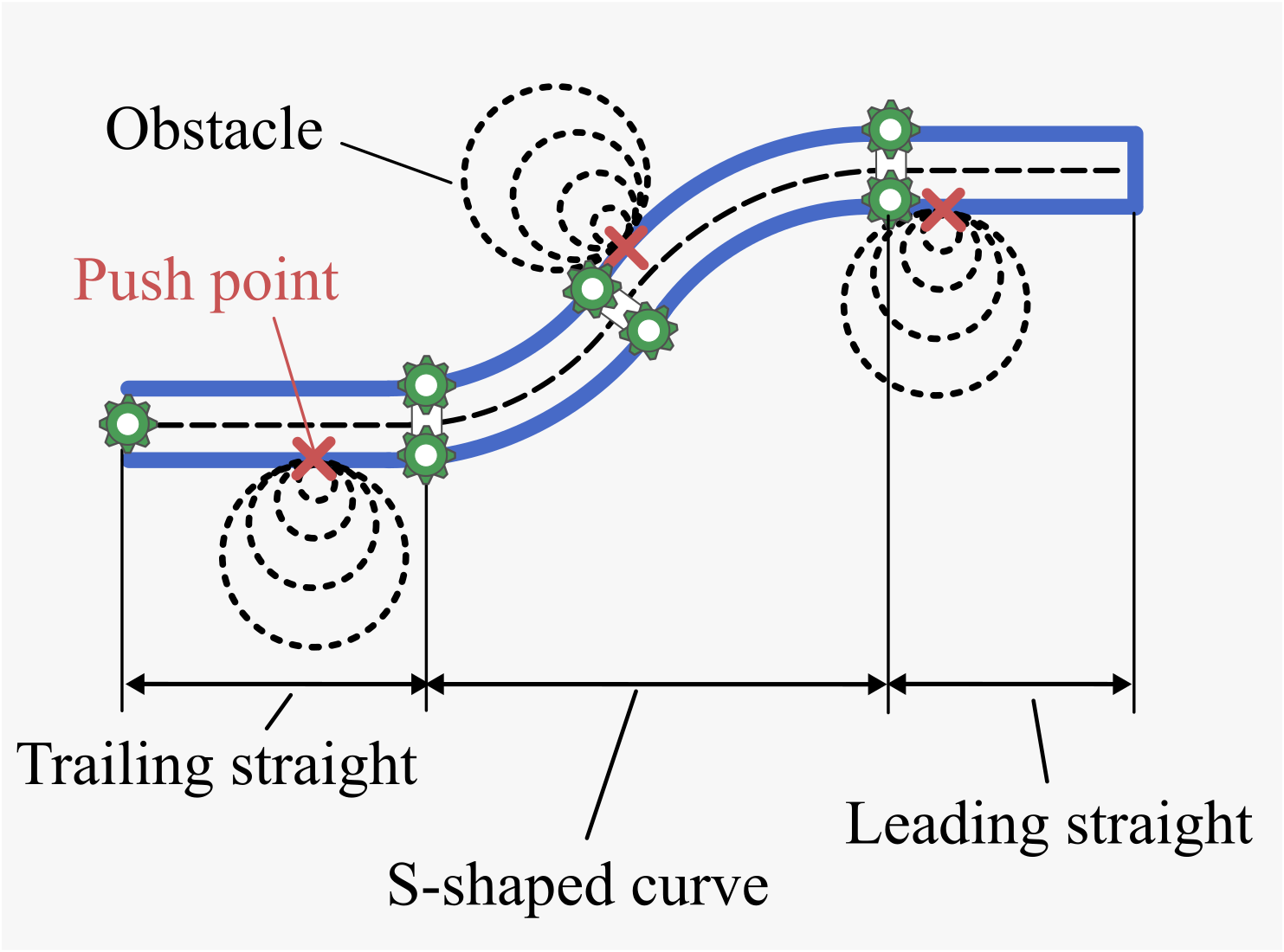}
  \caption{Target body shape composed of four segments and the geometric arrangement of the three alternating contact points.}
  \label{fig:target_shape}
\end{figure}

Specifically, the robot is configured with a total of four joint units: three movable joint units and one fixed joint unit. This number was selected because the target body shape is modeled as a sequence of ``Leading Line -- S-shaped Curve -- Trailing Line.'' Geometrically, the S-shaped curve is represented as a connection of two arcs with opposite curvature signs, and the straight lines can be regarded as arcs with zero curvature. Thus, the entire target shape is modeled as four continuous arc segments pressing against the three alternating obstacles (Fig.~\ref{fig:target_shape}).
Since the proposed mechanism requires one joint unit to determine the shape of one arc segment, four joint units are necessary to independently control these four segments.

The motion sequence is as follows:
\begin{enumerate}
    \item \textit{Initial Shape Formation}:
    The four joint units are driven to form the aforementioned ``Line--S-shape--Line'' profile, and the robot is positioned so that the intermediate S-shaped part presses against three obstacles.
    \item \textit{Shift of S-shaped Curve (Propulsion)}:
    To propel while maintaining the contact state, the body shape is shifted from anterior to posterior without deformation. In this robot, by controlling the motor speeds to be constant, all movable joint units travel backward along the rack gears at a uniform velocity while maintaining the bending state, realizing smooth shape propagation.
\end{enumerate}

\begin{figure}[t]
  \centering
  \includegraphics[width=0.9\linewidth]{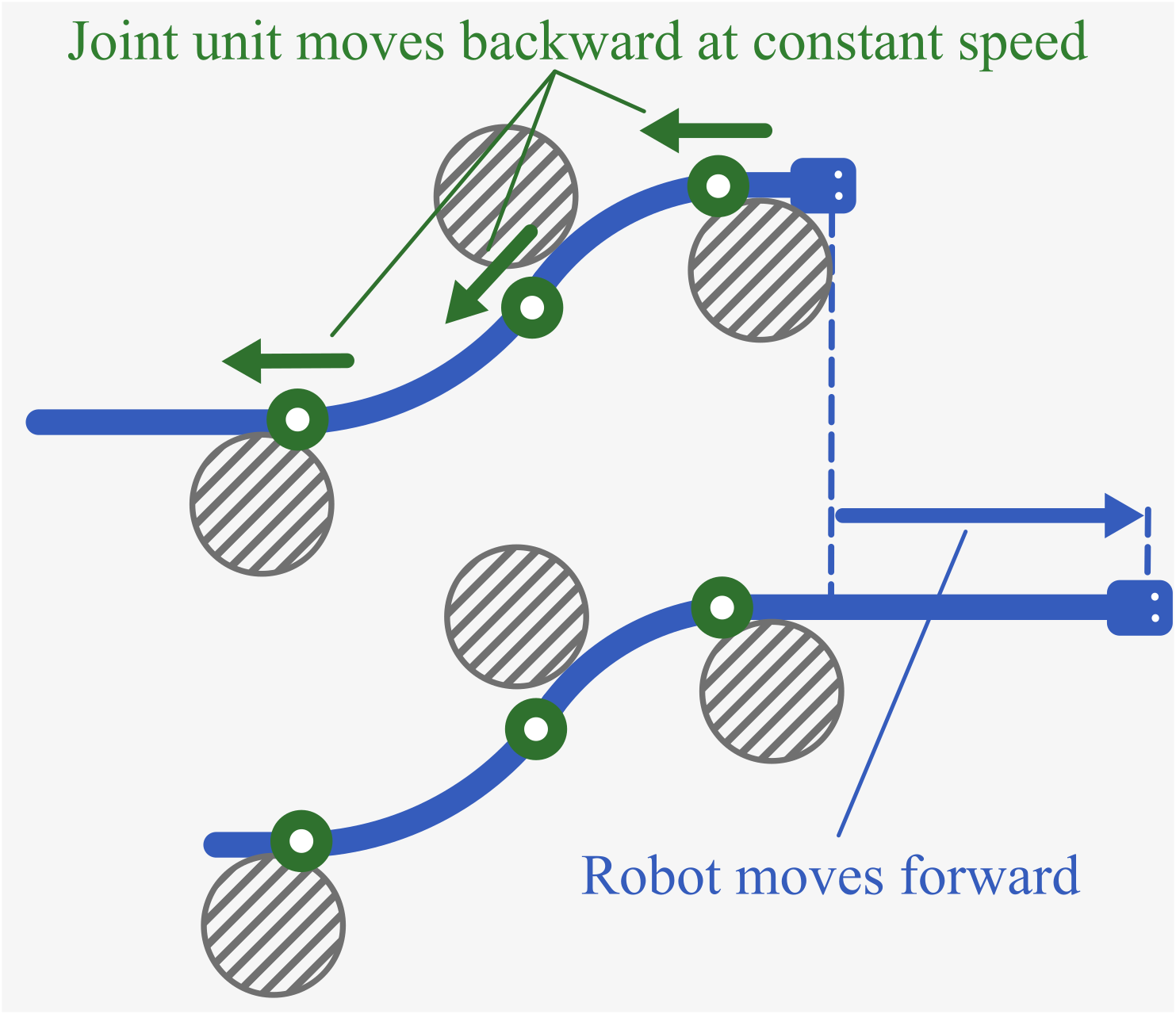}
  \caption{Obstacle-aided locomotion sequence via the continuous backward translation of the movable joint units.}
  \label{fig:OALmotion}
\end{figure}

Through this motion, the robot behaves as if a body wave is propagating backward, generating forward propulsion via reaction forces from the obstacles (Fig.~\ref{fig:OALmotion}).

\subsection{Control Parameters and Experimental Conditions}
Regarding propulsion control, both the High-Density and Low-Density models were given the exact same target shape. The S-shaped bending part consists of two smoothly connected arcs with a curvature radius $R = 420$~mm and a central angle $\theta = 25^\circ$ (opposite signs). The initial positions of the joint units were set such that the leading straight section was $100$~mm at the start of the experiment.
The Velocity Control Mode of the servomotors (Dynamixel XL430-W250-T, ROBOTIS) was used to drive the joint units. The target rotational speed was set to a constant $16$~rpm, and the operating time was $16$~s. Based on the active gear pitch radius ($r = 18$~mm) described in Section~II, the theoretical unit translation velocity is approximately $30.2$~mm/s at this rotational speed. Thus, all control parameters other than joint density are kept identical.

As the robot generates forward propulsion by translating the joint units backward, the movable joint units will eventually reach the rear end of the robot, preventing further forward movement. Therefore, to achieve continuous locomotion, a reset motion is necessary, in which the propulsion is temporarily halted, and the joint units return to the front while relaxing the body shape~\cite{kanada2025}. Since this study focuses on the effect of joint density on propulsive force generation, we evaluate only the propulsion phase (front-to-rear travel of the movable joint units) and exclude the reset motion.

%%%%%%%%%%%%%%%%%%%%%%%%%%%%%%%%%%%%%%%%%%%%%%%%%%%%%%%%%%%%%%%%%%%%%%%%%%%%%%%%
\section{EXPERIMENTAL EVALUATION}\label{sec:experiment}
In this section, we present comparative experiments using the developed High-Density and Low-Density models to verify the effect of joint density on the performance of obstacle-aided locomotion.

\begin{figure*}[t!]
  \centering
  \includegraphics[width=0.95\textwidth]{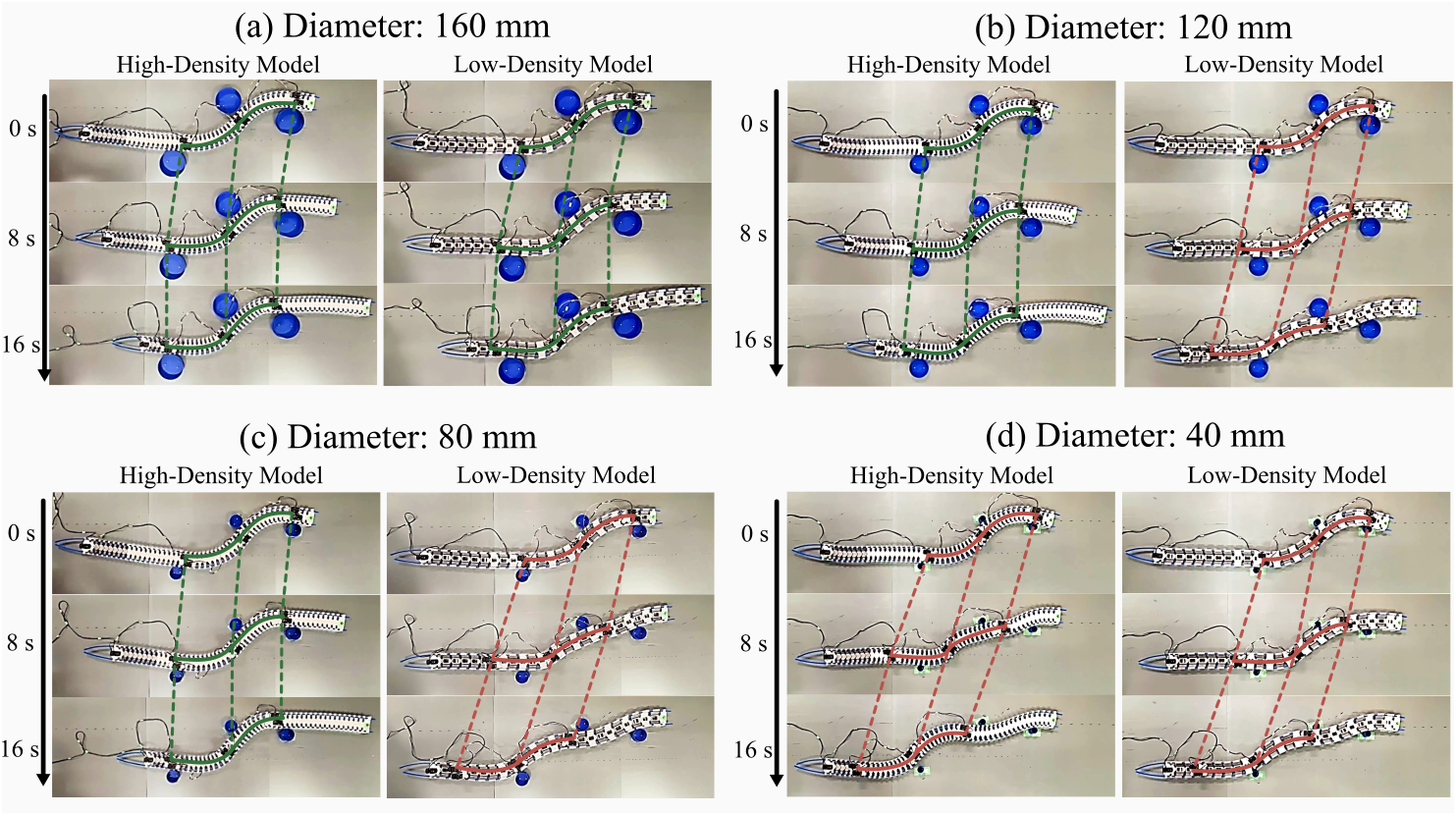}
  \caption{Snapshots of obstacle-aided locomotion: (a) $160$~mm diameter, (b) $120$~mm diameter, (c) $80$~mm diameter, and (d) $40$~mm diameter. The S-shaped bending sections are traced with solid lines. Green lines indicate successful propulsion, where the curve effectively maintains contact with the obstacles over time. Red lines indicate stagnation, where the curve fails to maintain geometric constraints from the obstacles and slips backward.}
  \label{fig:result_snap}
\end{figure*}

\subsection{Experimental Objectives}\label{subsec:experimental_purpose}
The objective of these experiments is to quantitatively demonstrate that stagnation (or jamming) caused by the discrete, link-by-link contact of conventional rigid-link snake robots can be mitigated by increasing joint density. In such robots, even a slight postural change induces discontinuous fluctuations in reaction forces, leading to stagnation. In contrast, the high-density structure aims to realize continuous contact and detachment across multiple links, suppressing these abrupt changes.

\subsection{Experimental Setup}\label{subsec:experimental_setup} 
\subsubsection{Obstacle Configurations} 
Cylindrical objects are used as obstacles. For obstacles with a smaller radius, the surface normal---and thus the reaction force vector---shifts more abruptly as the contact point moves along the link. This discontinuity causes propulsion loss and stagnation, making OAL more difficult.

To examine this effect, we prepared four obstacle environments with diameters of $160$, $120$, $80$, and $40$~mm. A smaller diameter yields a larger curvature relative to the link length, making contact more discrete and stagnation more likely. By varying the environment from relatively easy ($160$~mm) to highly challenging ($40$~mm) conditions, we quantitatively evaluate how well the High-Density model adapts to geometric difficulty.

Three obstacles were placed alternately on the left and right of the robot's initial shape. To ensure a fair comparison, the obstacle centers were not fixed; instead, as shown in Fig.~\ref{fig:target_shape}, they were positioned so that the contact points relative to the initial shape remained identical across environments.

A smooth steel plate was laid on the floor to eliminate leg-ground friction, so that propulsion arose solely from obstacle reaction forces. The 3D-printed obstacles were fixed to the plate with magnets and adhesive tape to prevent displacement during contact.

\subsubsection{Evaluation Metrics}
We used the following three metrics to evaluate the effect of joint density on propulsion performance.

\textit{Travel distance}: The position of a marker on the robot's head was tracked with a motion capture system, and the total travel distance was computed from the recorded coordinates.

\textit{Power consumption increment ($P_{\rm inc}$)}: When contact points transition into a state unfavorable for propulsion, the robot presses excessively against the obstacles instead of advancing, generating ineffective forces that appear as increased power consumption~\cite{Fu_2023}. To isolate this load, we used the increment $P_{\rm inc} = P_{\rm total} - P_{\rm base}$, where both terms are the power summed over all seven motors: $P_{\rm total}$ is measured under each condition, and $P_{\rm base}$ is the baseline during flat-ground locomotion without obstacles. The motor current and input voltage were acquired synchronously via the Dynamixel \textit{GroupSyncRead} command.

\textit{Cost of Transport (CoT)}: To assess energy efficiency independently of speed, we used the dimensionless $CoT = \bar{P}_{\rm total} / (mg\bar{v})$, where $m$ is the robot mass ($1.426$~kg for the High-Density model and $1.501$~kg for the Low-Density model), $g$ the gravitational acceleration, and $\bar{P}_{\rm total}$ and $\bar{v}$ the average power and forward velocity, respectively. The CoT was evaluated only during the active propulsion phase: if the forward velocity stayed below $0.5$~mm/s for $0.5$~s, that instant was taken as the onset of stagnation, and subsequent data---during postural collapse with reduced motor loads---were excluded to prevent underestimating the transport cost.

\begin{figure*}[t!]
  \centering
  \includegraphics[width=0.95\textwidth]{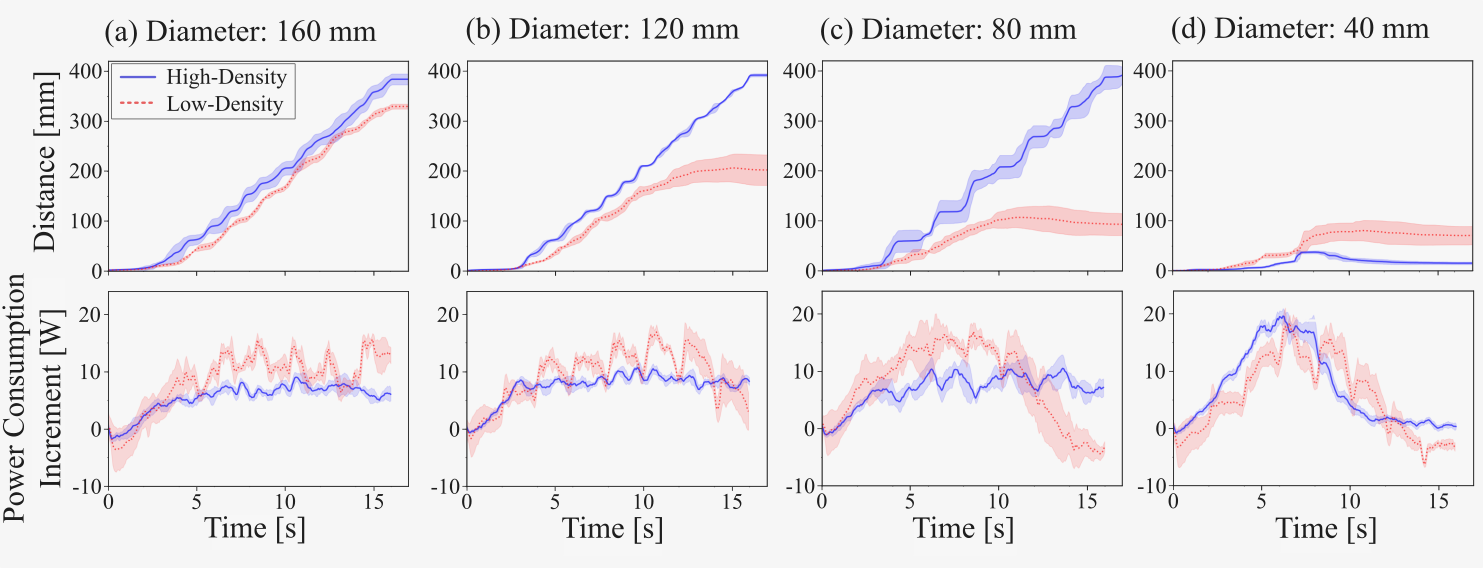}
  \caption{Time evolution of travel distance (upper) and power consumption increment $P_{\rm inc}$ (lower) in each obstacle environment. The solid lines and shaded regions indicate the mean values and standard deviations across four trials, respectively.}
  \label{fig:result_distance_and_power}
\end{figure*}

\subsection{Results}\label{subsec:results}
% --- Postural Behavior ---
The overall differences in postural behavior between successful propulsion and stagnation can be visually observed in the snapshots shown in Fig.~\ref{fig:result_snap}. In conditions where propulsion was maintained, the S-shaped curvature of the robot's body effectively maintained the contact points formed with the obstacles, pushing the body forward. Conversely, in conditions that led to stagnation, the S-shaped curvature failed to maintain these contact points. Rather than simply losing its shape, the bending section slipped backward relative to the obstacles, resulting in a loss of forward propulsion.

% --- Travel Distance ---
To quantitatively evaluate these locomotion behaviors, the upper plots in Fig.~\ref{fig:result_distance_and_power} show the time evolution of the total travel distance for the High-Density and Low-Density models across each obstacle environment. The High-Density model maintained continuous propulsion in all environments except for the $40$~mm diameter case. In contrast, while the Low-Density model was able to continue its propulsive motion in the $160$~mm environment, its final travel distance was approximately $54$~mm shorter than that of the High-Density model. In the other small-diameter environments, the travel distance of the Low-Density model ceased to increase midway through the motion, and the robot was observed to exhibit stagnation (i.e., the joints continued to actuate, but the body as a whole failed to advance forward).

% --- Power Consumption ---
The lower plots in Fig.~\ref{fig:result_distance_and_power} show the time evolution of the power consumption increment ($P_{\rm inc}$) in each environment. In the $160$~mm environment where both models succeeded, the High-Density model suppressed the power increase caused by obstacle interaction more effectively than the Low-Density model (High-Density: approx. $5.4$~W avg., Low-Density: approx. $8.5$~W avg.). Throughout the periods when both models were moving, the High-Density model generally exhibited lower power consumption. In the $40$~mm environment where both models failed, the power consumption peaked at the onset of stagnation (High-Density: approx. $19.6$~W, Low-Density: approx. $18.3$~W) before decreasing as the posture collapsed.

% --- CoT ---
\begin{figure}[tb]
  \centering
  \includegraphics[width=0.9\linewidth]{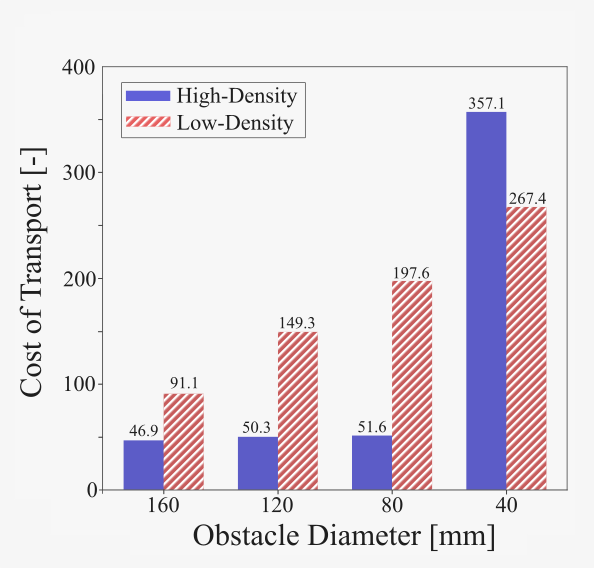}
  \caption{Comparison of the Cost of Transport (CoT). Lower values indicate higher efficiency. For stagnated trials (both models at $40$~mm; Low-Density at $120$ and $80$~mm), CoT was calculated only from the active phase before complete stagnation.}
  \label{fig:cot_bar}
\end{figure}

Furthermore, to comprehensively evaluate the energy efficiency reflecting both the variation in forward velocity and power consumption, Fig.~\ref{fig:cot_bar} visualizes the Cost of Transport (CoT) during the active propulsion phase for each environment.

As shown in Fig.~\ref{fig:cot_bar}, the High-Density model achieved consistently high energy efficiency (low CoT, remaining around $47$--$52$) across the successful environments ($160$~mm, $120$~mm, and $80$~mm). In stark contrast, the CoT of the Low-Density model increased markedly from $91.1$ ($160$~mm) to $197.6$ ($80$~mm) as the obstacle diameter decreased. This indicates that the low-density configuration suffers from discontinuous contacts that drastically drop the forward velocity, expending excessive energy on non-propulsive forces per unit distance traveled. In the $40$~mm environment, both models effectively failed to propel, resulting in extremely high CoT values due to near-zero velocities. These quantitative metrics clearly demonstrate that a high joint density is effective in suppressing abrupt shifts in contact configurations and converting environmental constraints into stable, efficient propulsion.

\subsection{Discussion}\label{subsec:discussion}
The observed differences in propulsion performance and energy efficiency (CoT) can be explained by two key geometric factors: the continuity of contact transitions and the structural gaps between links.

First, for efficient obstacle-aided locomotion, reaction forces must be handed over seamlessly between links. In the Low-Density model, long rigid links create large angular deviations before detachment, forcing the next link to initiate contact at a suboptimal angle. This discontinuous transition directs reaction forces inward as lateral loads, deforming the robot's body and wasting excessive motor power, which drastically increases the CoT. Conversely, the High-Density model approximates a continuous curve, smoothing these angular deviations to maintain efficient forward propulsion across the $160$--$80$~mm environments.

Second, despite the advantage of high joint density, both models ultimately stagnated in the $40$~mm environment due to the inter-link gaps required for joint rotation. While these gaps only caused minor periodic velocity drops (pulsation) in larger environments, the $40$~mm obstacles were small enough to lodge directly into them, inducing complete geometric locking. Because the motors continued attempting to track target angles against the immovable obstacles, the body posture collapsed, resulting in the sharp power peaks and extreme CoT values.

%%%%%%%%%%%%%%%%%%%%%%%%%%%%%%%%%%%%%%%%%%%%%%%%%%%%%%%%%%%%%%%%%%%%%%%%%%%%%%%%
\section{CONCLUSION}\label{sec:conclusion} 
This study addressed the problem of stagnation caused by the discrete contact of rigid links in snake robot obstacle-aided locomotion. We proposed a structural approach through a ``high-density joint configuration'' as a solution. To verify the effectiveness of this approach, we developed two models with different joint densities and conducted comparative experiments in environments with varying levels of geometric difficulty (obstacle diameters).

The experimental results demonstrated that high-density joints allow for the maintenance of smooth contact with obstacles, suppressing the generation of ineffective internal forces and energy loss. This effect was particularly pronounced in challenging environments with small-diameter obstacles. The Low-Density model suffered from postural deformation and stagnation due to geometric incompatibility. In contrast, the High-Density model maintained appropriate contact points, achieving stable and efficient propulsion. Based on these findings, we conclude that high-density joint configurations provide a key design guideline for enabling snake robots to adapt to complex, irregular environments through stable obstacle-aided locomotion.

Future work will focus on three key areas. First, we aim to implement continuous long-distance locomotion based on our proposed obstacle-aided locomotion strategy. Second, we plan to introduce autonomous adaptive control by measuring the reaction force through pressure sensing on the robot's sides. Finally, the hardware could be improved by covering the body with a flexible skin to eliminate the inter-link gaps, thereby preventing geometric locking and further enhancing traversal performance in extreme environments.

\addtolength{\textheight}{-3cm}   % This command serves to balance the column lengths

%%%%%%%%%%%%%%%%%%%%

%%%%%%%%%%%%%%%%%%%%%%%%%%%%%%%%%%%%%%%%%%%%%%%%%%%%%%%%%%%%

\bibliographystyle{IEEEtran}
\bibliography{reference}

\end{document}